\documentclass[conference]{IEEEtran}
\IEEEoverridecommandlockouts
\usepackage{cite}
\usepackage{amsmath,amssymb,amsfonts}
\usepackage{graphicx}
\usepackage{textcomp}
\usepackage{subfigure}
\usepackage{algorithm}
\usepackage[noend]{algpseudocode}
\usepackage[normalem]{ulem}
\usepackage{fancyhdr}
\usepackage{dsfont}
\usepackage{comment}
\usepackage{enumitem}
\usepackage{url}
\usepackage[lmargin=0.7in, rmargin=0.7in, tmargin=0.7in, bmargin=1in]{geometry}
\usepackage{multirow}
\usepackage{rotating}
\usepackage[table,xcdraw]{xcolor}

\newcommand\blfootnote[1]{%
  \begingroup
  \renewcommand\thefootnote{}\footnote{#1}%
  \addtocounter{footnote}{-1}%
  \endgroup
}

\def\BibTeX{{\rm B\kern-.05em{\sc i\kern-.025em b}\kern-.08em
    T\kern-.1667em\lower.7ex\hbox{E}\kern-.125emX}}
\begin{document}

\title{
The Robustness of Spiking Neural Networks in Communication and its Application towards Network Efficiency in Federated Learning
}


 \author{\IEEEauthorblockN{Manh V. Nguyen\IEEEauthorrefmark{1},
 Liang Zhao\IEEEauthorrefmark{1},
 Bobin Deng\IEEEauthorrefmark{1},
 William Severa\IEEEauthorrefmark{2},
 Honghui Xu\IEEEauthorrefmark{1},
 Shaoen Wu\IEEEauthorrefmark{1}}
 \IEEEauthorblockA{\IEEEauthorrefmark{1}College of Computing and Software Engineering,
 Kennesaw State University, Georgia, USA
 \\mnguy126@students.kennesaw.edu, \{lzhao10, bdeng2, hxu10, swu10\}@kennesaw.edu}
 \IEEEauthorblockA{\IEEEauthorrefmark{2}Department of Cognitive \& Emerging Computing,
 Sandia National Laboratories, New Mexico, USA
 \\wmsever@sandia.gov}
 }

\maketitle 

\blfootnote{This paper has been accepted for publication at the 43\textsuperscript{rd} IEEE International Performance Computing and Communications Conference (IPCCC 2024).}

\begin{abstract}
Spiking Neural Networks (SNNs) have recently gained significant interest in on-chip learning in embedded devices and emerged as an energy-efficient alternative to conventional Artificial Neural Networks (ANNs).
However, to extend SNNs to a Federated Learning (FL) setting involving collaborative model training, the communication between the local devices and the remote server remains the bottleneck, which is often restricted and costly. 
In this paper, we first explore the inherent robustness of SNNs under noisy communication in FL. Building upon this foundation, we propose a novel Federated Learning with Top-$\kappa$ Sparsification (FLTS) algorithm to reduce the bandwidth usage for FL training. 
We discover that the proposed scheme with SNNs allows more bandwidth savings compared to ANNs without impacting the model's accuracy. 
Additionally, the number of parameters to be communicated can be reduced to as low as $6\%$ of the size of the original model.
We further improve the communication efficiency by enabling dynamic parameter compression during model training. 
Extensive experiment results demonstrate that our proposed algorithms significantly outperform the baselines in terms of communication cost and model accuracy and are promising for practical network-efficient FL with SNNs.



\end{abstract}


\section{Introduction}
Spiking Neural Networks (SNNs) constitute a significant advancement in artificial intelligence (AI), operating with asynchronous discrete events called spikes, which enhances energy efficiency, particularly for neuromorphic hardware designed to mimic the brain's neural architecture. 
This efficiency is crucial as AI's energy demands rise, positioning SNNs as a solution to the impending energy crisis \cite{mark_energy_2024}. 
Beyond neuromorphic hardware, SNNs are also suitable for general-purpose use, including on resource-constrained edge devices with limited power budgets. 
Studies and practical applications have demonstrated that they provide significant energy and resource efficiency~\cite{dampfhoffer2022snns,kundu2024recent}, making SNNs an attractive option for sustainable, low-power AI systems across various domains, from industrial automation to smart home technologies.

For efficiency and privacy concerns, Federated Learning (FL) has emerged as an effective framework for distributed AI systems. It has great potential to be integrated with SNNs \cite{venkatesha_federated_2021} to enjoy the benefits of both worlds. With FL approaches, instead of aggregating the raw data from distributed entities to a central server, the model is trained across multiple devices, each holding local data samples. The central server then aggregates the updates of the locally trained models. 
However, FL training methods also incur significant communication costs due to frequent updates and exchanges of model parameters between the central server and the decentralized nodes. This makes communication the bottleneck in the large-scale deployment of FL systems. 
Additionally, most existing FL frameworks
assume perfect communication while
real-world communication networks are noisy and prone to packet loss and transmission errors \cite{Ye2022_FL_unreliablecomm}. Noisy communication can lead to corrupted data exchanges, negatively impacting the model's convergence and accuracy. 
Addressing these issues is crucial for the widespread adoption and efficiency of FL systems.

This work attempts to improve network efficiency in FL with SNNs considering noisy communication channels. 
We first explore the robustness of SNNs compared to Artificial Neural Networks (ANNs) as a foundation for its adaptation in bandwidth-limited environments. By leveraging the robustness of SNNs to noise, we propose a suite of Top-$\kappa$ Sparsification based algorithms to reduce bandwidth consumption in FL with SNN.
Finally, we conduct empirical experiments to verify the superiority of SNNs to ANNs in FL settings under noisy communication scenarios. Furthermore, we evaluate the effectiveness of the proposed algorithms in terms of communication cost and accuracy and their sensitivity to network size and communication compression rate. 
This work aims to provide practical and advantageous FL solutions for edge-device SNNs, whose environments are often under limited and noisy communication settings.

\textbf{Contribution.} The key contribution of the work presented in this paper is three-fold:
\begin{itemize}[leftmargin=*]
    \item To the best of our knowledge, we are among the first to investigate the performance of FL with SNN under noisy communication. We discover that SNN is significantly more robust than equivalent ANN models.
    \item In light of the inherent robustness of SNN, we propose Federated Learning with Top-$\kappa$ Sparsification to improve communication efficiency. Additionally, we leverage the principle of critical learning periods in Federated Learning and propose a Federated Learning with Dynamic-$\kappa$ Reduction to further reduce the communication overhead. 
    \item We conduct extensive experiments to validate the proposed algorithms. 
    The results demonstrate that the novel algorithms enable SNN to save significant bandwidth compared to ANN in FL settings with noisy communication. Further, they can achieve comparable model accuracy with $6\%$ of the communication cost compared to the baselines. 
    
     
\end{itemize}


\section{Background of Federated Learning and Spiking Neural Networks} \label{sec:background}

In this section, we briefly cover the background of Spiking Neural Networks (SNNs) and Federated Learning (FL). 

\subsection{Spiking Neural Network}

\begin{figure}
    \centering
    \includegraphics[scale=0.65]{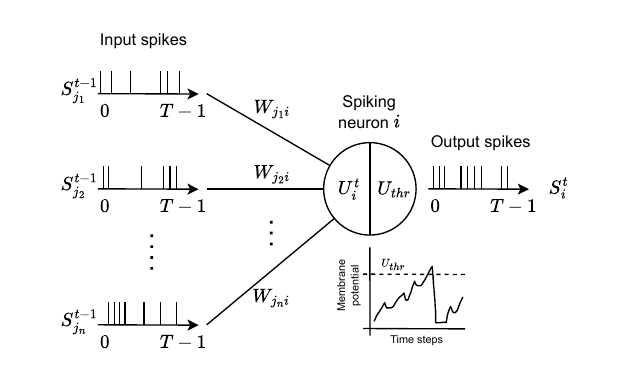}
    \caption{Processing Mechanism of Integrate-and-Fire (IF) Spiking Neuron.}
    \label{fig:snn_diagram}
\end{figure}

\textbf{Rate encoding}. A spiking neuron is modeled to perceive input as incoming spikes over a predefined time interval. The neuron accumulates membrane potential while absorbing the input spikes and scales by the synaptic weights. Once the membrane potential reaches a certain threshold, the neuron fires an output spike, releases the membrane energy, and the process restarts. This mechanism is known as Integrate-and-Fire (IF) and is illustrated in Fig \ref{fig:snn_diagram}. Another popular type of spiking neuron is  Leaky-Integrate-and-Fire (LIF), which gradually decreases (or leaks) membrane potential at each timestep. The following equation describes the LIF mechanism of a neuron $i$:

\begin{multline}
U_i^t = \sum_{j \in N} W_{ij} S_{j}^{t-1} + \beta U_i^{t-1} - S_{i}^{t-1} U_{thr},
\\
\text{where } \beta < 1 \text{ and } S_i^{t} = \begin{cases} 
1 & \text{if } U_i^{t} > U_{thr}, \\
0 & \text{otherwise}. \end{cases}
\end{multline}
in which, $S_i^t$ is the binary output of neuron $i$, and $U_i^t$ is its membrane potential at timestep $t$. 
While $\beta$ represents the leak factor by which the membrane potential is reduced at each timestep and $U_{thr}$ represents the membrane threshold. $N$ denotes the set of input neurons that is connected to $i$, and $W_{ji}$ denotes the synaptic weight of the $j\rightarrow i$ connection. 



\textbf{Back-propagation method for model training}. Due to the time dimension encoding and the non-differentiable functions, the conventional ANN backpropagation is unsuitable for SNN training. According to Neftci et al.\cite{neftci2019surrogate}, the back-propagation of a spiking neuron begins by calculating $\Delta W_{ji}$, the gradient of the weight connecting neuron $i$ and $j$ accumulated over $T$ as follows:
\begin{equation}
    \Delta W_{ji} = \sum_{t=1}^{T} \frac{\partial \mathcal{L}}{\partial S_i^t} \frac{\partial S_i^t}{\partial U_i^t} \frac{\partial U_i^t}{\partial W_{ji}},
\end{equation}
where $\mathcal{L}$ is the loss function. Categorical cross-entropy is widely used in image classification. As $S_i^t$ is a thresholding function, computing $\Delta W_{ji}$ is intractable. To simplify this problem, the threshold function can be approximated with surrogate functions that is defined as follows:
\begin{equation}
    \frac{\partial S_i^t}{\partial U_i^t} = \xi \max \left\{ 0, 1 - \left| \frac{U_i^t - U_{thr}}{U_{thr}} \right| \right\},
\end{equation}
where $\xi$ is a hyperparameter decay factor for back-propagated gradients\cite{neftci2019surrogate,bohte2011error}. As gradients are accrued through time, $\xi$ is adjusted according to $T$. The larger the timesteps, the smaller the decay factor should be to avoid gradient exploding. 

\subsection{Federated Learning}

FL framework typically consists of one global aggregation server and a set of local training clients $\mathcal{C}$, in which each client $c$ possesses its private dataset $\mathcal{D}^{(c)}$. The server starts the training with a global initial model weight $\mathbf{W}_0$ and is distributed to the clients. For each round $r$, client $k$ receives the global updated model $\mathbf{W}_{r-1}$ aggregated from the previous round. Then, each client uses its private data samples for local training and obtains the locally updated model $\mathbf{W}_r^{(c)}$. The client transfers its updated parameters to the server for aggregation and produce the global model $\mathbf{W}_r$, the following formula describes this \textit{FedAvg} algorithm \cite{mcmahan2017communication}: 
\begin{equation}
    \mathbf{W}_{i} = \frac{1}{|\mathcal{C}|} \sum_{c \in \mathcal{C}} \mathbf{W}_i^{(c)} ,
\end{equation}
in which, $|\mathcal{C}|$ denotes the number of participated clients in the FL network. Note that there is a rich literature on alternatives for model aggregation algorithms. For simplicity, this work will adopt the \textit{FedAvg} algorithm for evaluation without loss of generality. Fig. \ref{fig:fl_diagram} illustrates an overview of FL with SNN integrated with the compression schemes that we are proposing in this work. 

\subsection{SNN In-situ Training Hardware/System}
This paper investigates on top of the client-server architecture in which the server or each client has SNN in-situ training capability. Many recent research works~\cite{Li2023, Lan2023,vohra2024circuit} focus on SNN in-situ training and have made significant progress. We expect that in-situ training will become an essential feature of neuromorphic hardware/systems in the near future. 

\section{Robustness of SNN in FL with noisy communication} \label{sec:robust}

Recently, Patel et al. \cite{patel_impact_2023} investigated the impact of noisy input on the SNN training phase and showed that SNN models can be more resilient than their ANN counterparts. This observation motivates us to explore the robustness of SNN to practical noisy communication in FL, which is an under-explored research area. In this section, we extend SNNs to the FL setting under noisy communication and compare the noise robustness of SNNs with ANNs.




\textbf{Setup}. To compare the noise robustness between SNNs and ANNs in a FL environment, we utilize VGG9 as the base architecture for both models, with stochastic gradient descent (SGD) optimizer and average pooling. We follow the setup in ~\cite{venkatesha_federated_2021} and adopt similar training parameters. For SNN, we set the timesteps to $25$, with a learning rate of $0.1$ and momentum of $0.95$, while for ANN, we set the learning rate of $0.001$ and weight decay of $5e-4$. We set up the federated learning system with $5$ clients, with a batch size of $32$. We train for $40$ global aggregation rounds. Before aggregating each global aggregation, each client trains on its private data for $5$ local epochs. The model noise is added before transferring global parameters from server to client and locally trained parameters from client to server. The noise is generated following the Gaussian distribution $N(0, \sigma)$, in which the standard deviation, i.e., $\sigma$, indicates the strength. 



\begin{figure}
    \centering
    \subfigure[]{\includegraphics[scale=0.4]{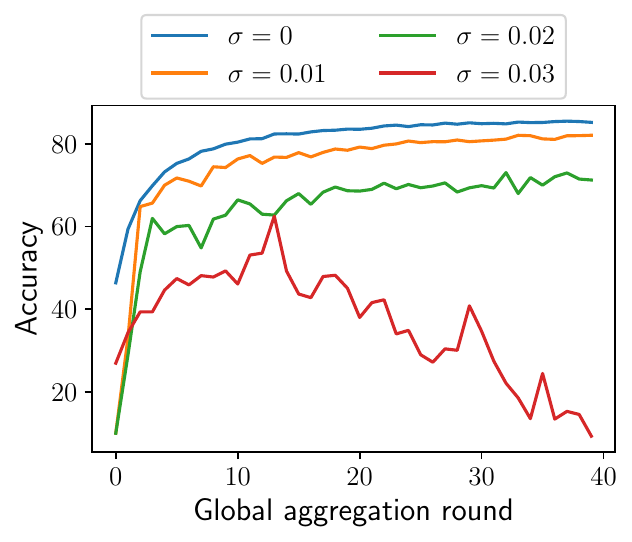}\label{fig:acc_v_noise_a}}
    \subfigure[]{\includegraphics[scale=0.4]{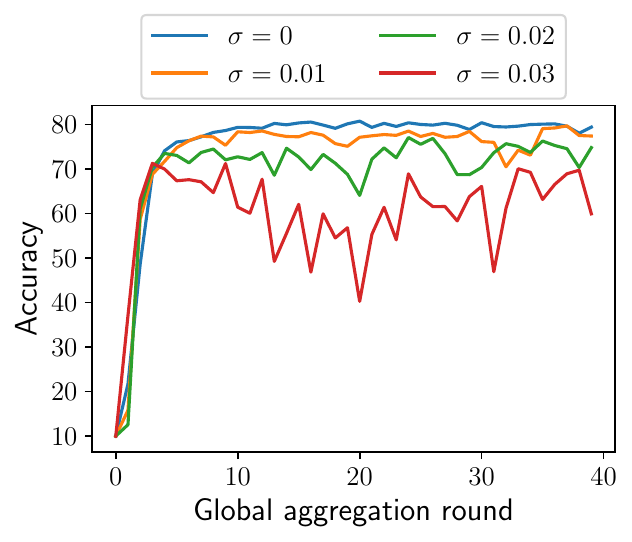}\label{fig:acc_v_noise_b}}
    \caption{The stability of a) ANN and b) SNN under various noise levels $\sigma$ using absolute values.}
    \label{fig:acc_v_noise}
\end{figure}

\begin{figure}
    \centering
    \subfigure[]{\includegraphics[scale=0.4]{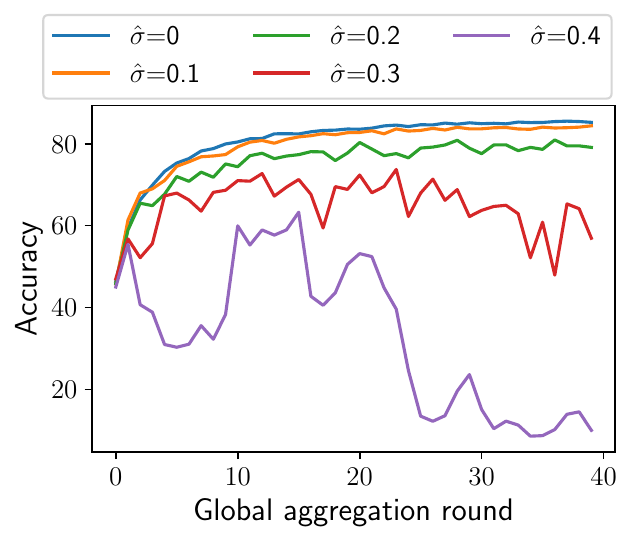}}
    \subfigure[]{\includegraphics[scale=0.4]{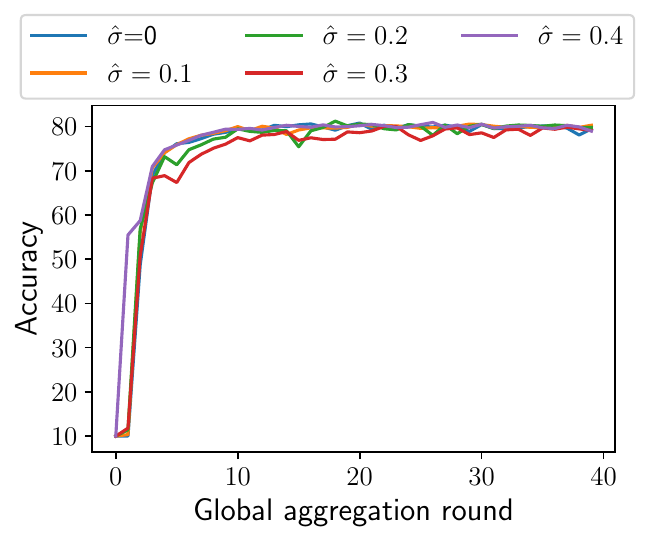}}
    \caption{The stability of a) ANN and b) SNN under various noise levels $\hat{\sigma}$ using relative percentage to the magnitude of the parameters.}
    \label{fig:acc_v_rltv_noise}
\end{figure}

\textbf{Noise generation}. In our evaluation, we vary the noise strength and compare the accuracy stability to verify the model robustness. We conduct experiments for ANN and SNN models by setting the noise strength. First, $\sigma$ is set as a fixed value from round to round, called absolute noise (Fig. \ref{fig:acc_v_noise}). Second, $\sigma$ is fractionally dependent on the average magnitude of the transmitting parameters at each round, called relative noise (Fig. \ref{fig:acc_v_rltv_noise}). With $\hat{\sigma}$ to annotate the relative noise strength, we have the following noise generation scheme:
\begin{equation}
    \sigma_r = \hat{\sigma} \times \textsc{Avg} \left ( \left \{ |\mathbf{w}| : \forall \mathbf{w} \in \mathbf{W}_r \right \} \right ),
\end{equation}
in which $\sigma_r$ is the strength of the noise added to the parameters set, $\mathbf{W}_r$, when transmitting at round $r$. 

\textbf{Robustness comparison for ANN and SNN}. In Fig.~\ref{fig:acc_v_noise}, three different absolute noise levels $\sigma \in \{0.01,0.02,0.03\}$ are compared with the no-noise baseline in both ANN (shown in Fig.~\ref{fig:acc_v_noise_a}) and SNN (shown in Fig.~\ref{fig:acc_v_noise_b}). According to our experimental results, even though absolute noise introduces instability to both models, the noise at $\sigma = 0.03$ causes the ANN model to be more saturated over time, while the SNN model is still robust enough to maintain relatively stable accuracy. On the other hand, in Fig. \ref{fig:acc_v_rltv_noise}, we also compare the no-noise baseline of both models with four different relative noise levels $\hat{\sigma} \in \{0.1, 0.2, 0.3, 0.4\}$. 
As the relative noise strength increases, the accuracy of ANN drops dramatically and becomes unacceptable. However, the accuracy of SNN remains stable with all these noise levels. Therefore, we conclude that SNNs consistently outperform ANNs in terms of robustness in FL under noisy communication. 


\section{Proposed Communication-efficient FL Algorithms with SNN} \label{sec:propose}

\subsection{Rationale}
\label{subsec:alg-motivation}

\begin{figure}
    \centering
    \includegraphics[width=0.9\linewidth]{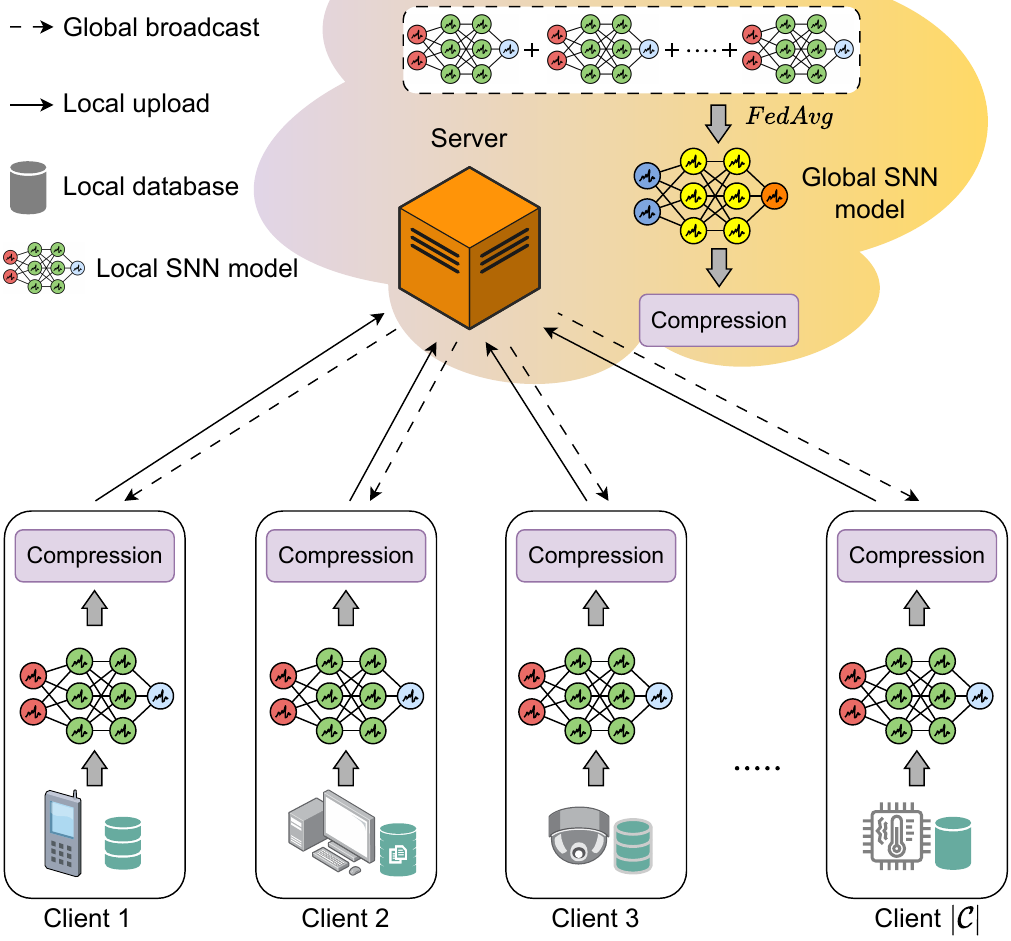}
    \caption{Overview of Federated Learning with Spiking Neural Networks integrated with the proposed model compression algorithms.}
    \label{fig:fl_diagram}
\end{figure}

As explored in Section \ref{sec:robust}, the property of robustness to noise demonstrated by SNN implies its potential resilience to other modes of manipulation of the parameters being transferred between FL clients and the server.
In light of this observation, we are motivated to propose a FL algorithm with SNN that aims to reduce the communication load among the nodes while avoiding compromises in the quality of the trained model. 
Fig. \ref{fig:fl_diagram} shows the integration of the compression algorithms into the communication flow of FL with SNNs.


\subsection{Federated Learning with Top-$\kappa$ Sparsification}
 \label{subsec:FLTS}
\begin{algorithm}[t]
\caption{FL with Top-$\kappa$ Sparsification (FLTS)}\label{alg:adaptive_compression}
\label{alg:FLTS}
\textbf{Input:} $\mathcal{C}$, $\{\mathcal{D}^{(c)}: c \in \mathcal{C}\}$, $R$, $\kappa$

\textbf{Output:} $\mathbf{W}_R$
\begin{algorithmic}[1] 
\State Initialize $\mathbf{W}_0$
\State $j \gets 0$

\For{$r = 1$ to $R$} \Comment{Global training loop}

    \noindent \textit{Stage 1: Distribute global parameters}
    \State $\mathbf{W}^\prime_{r-1} \gets  \textsc{Sparse}(\mathbf{W}_{r-1}, \mathbf{H}_{r-1}, \kappa)$, equation \eqref{sparse-function}
    \State Distribute $\mathbf{W}^\prime_{r-1}$ to clients

    \noindent \textit{Stage 2: Federated training (Client-side)}
    \For{each client $c \in \mathcal{C}$}
        \State Train model $\mathbf{W}^{(c)}_r$ with local dataset $\mathcal{D}^{(c)}$
        \State $\mathbf{W}^{\prime (c)}_r \gets \textsc{Sparse}(\mathbf{W}^{(c)}_r, \mathbf{H}^{(c)}_r, \kappa)$
        \State Submit $\mathbf{W}^{\prime (c)}_r$ to the server
    \EndFor 

    \noindent \textit{Stage 3: Model aggregation}
    \vspace{3pt}
    \State $\mathbf{W}_{r} \gets \textsc{FedAvg} \left ( \left \{\mathbf{W}^{\prime (c)}_r: c \in \mathcal{C} \right\} \right )$
\EndFor 
\State \textbf{return} $\mathbf{W}_R$
\end{algorithmic}
\end{algorithm}

We propose Federated Learning with Top-$\kappa$ Sparsification (FLTS) leveraging the robustness of SNNs to improve communication efficiency. 
The details are described in Algorithm \ref{alg:FLTS} in stages. At \textit{Stage 1}, the updated model generated from the previous global aggregation is broadcast to all participating clients. At \textit{Stage 2}, clients perform independent training by utilizing private data samples and submitting local gradients to the server. At \textit{Stage 3}, the server performs the \textit{FedAvg} algorithm to aggregate the data gathered from clients and generate a new global model. For the transmissions between clients and the server in \textit{Stage 1} and \textit{Stage 2}, we design function $\textsc{Sparse}$ to implement our Top-$\kappa$ Sparsification scheme as follows. The goal is to reduce the transferring load and, therefore, to lower the total bandwidth overhead in the FL process. 


\textbf{Top-$\kappa$ Sparsification}. 
Our proposed Top-$\kappa$ parameters sparsification scheme is a compression protocol, which modifies the sparsification scheme in \cite{zheng2023reducing}. The goal is to transmit only top parameters with the largest gradients (absolute values) in the resulting model. The insight of this approach is that the magnitudes of parameter gradients are associated with their importance to the model update. Thus, we can potentially preserve the important information in communication by transferring only the model updates with large magnitudes. Specifically, assume that $\mathbf{W} \in \mathds{R}^d$ is the trained model, and the corresponding gradient of this model is $\mathbf{H} \in \mathds{R}^d$. Mathematically, the gradient of a client model after local training of round $r$ is  $\mathbf{H}^{(c)}_r = \mathbf{W}^{(c)}_r - \mathbf{W}_{r - 1}$. In the same manner, the gradient of the global model after aggregation is $\mathbf{H}_r = \mathbf{W}_r - \mathbf{W}_{r - 1}$. In this work, we define a variable $\kappa \leq 1$ for the compression rate, i.e., the fraction of preserved parameters. Such that the number of parameters to be preserved is $u= \kappa |\mathbf{W}|$, in which $|\mathbf{W}|$ is the total number of parameters. The Top-$\kappa$ Sparsification function is defined as:
\begin{equation}\label{sparse-function}
    \textsc{Sparse}(\mathbf{W}, \mathbf{H}, \kappa) = \left\{ \mathbf{w}_{i_1}, \mathbf{w}_{i_2}, ..., \mathbf{w}_{i_u} \right\},
\end{equation}
where $i_1, i_2, ..., i_u \leq |\mathbf{W}|$ are parameter indices and  $\mathbf{h}_{i_1}, \mathbf{h}_{i_2}, ..., \mathbf{h}_{i_u} \in \mathbf{H}$ are gradients with top absolute values.


\subsection{Federated Learning with Dynamic-$\kappa$ Reduction} \label{subsec:FLDR}

\begin{algorithm}[t]
\caption{FL with Dynamic-$\kappa$ Reduction (FLDR)}\label{alg:FLDR}
\textbf{Input:} $\mathcal{C}$, $\{\mathcal{D}^{(c)}: c \in \mathcal{C}\}$, $R$, $(\alpha, \omega)$

\textbf{Output:} $\mathbf{W}_R$
\begin{algorithmic}[1] 
\State Initialize $\mathbf{W}_0$
\State $\kappa \gets \alpha$ \Comment{Initialize $\kappa$ at the highest value}

\For{$r = 1$ to $R$}
    
    \noindent \textit{Stages 1, 2 \& 3: Execute as Algorithm \ref{alg:FLTS} with current $\kappa$}

    \noindent \textit{Stage 4 (new): Reduce $\kappa$ value}
    \State $\kappa \gets \textsc{Reduce}(\kappa, \alpha, \omega, R)$, equation \eqref{fldr-l} or \eqref{fldr-e}
\EndFor 
\State \textbf{return} $\mathbf{W}_R$
\end{algorithmic}
\end{algorithm}



Notice that for FLTS in the last section, the compression rate $\kappa$ is constant throughout the training process. 
However, as pointed out by Yan et al. in \cite{yan2022seizing, yan2023criticalfl}, as the global model is gradually being developed through the global aggregation in FL training, the number of important parameters needed to be passed around can be gradually decreased. 
Specifically, we can potentially further reduce the communication cost by dynamically adjusting the compression rate $\kappa$. 
Therefore, in this section, we design a modified FLTS called Federated Learning with Dynamic-$\kappa$ Reduction (FLDR) to further reduce the total bandwidth consumption for FL with SNNs. 
The algorithm is presented in Algorithm. \ref{alg:FLDR}, where $\kappa$ is dynamically adjusted in each global aggregation round. We describe the heuristic $\kappa$ selection strategy as follows.

\textbf{Dynamic-$\kappa$ reduction}.
The adjustment of $\kappa$ is implemented through the $\textsc{Reduce}(\kappa, \alpha, \omega, R)$ function. 
We implement two modes of $\kappa$ reduction. The first mode relies on linear reduction, in which the function is formulated as follows:
\begin{equation}\label{fldr-l}
    \textsc{Reduce}(\kappa, \alpha, \omega, R) = \kappa - \frac{\alpha - \omega}{R}.    
\end{equation}
The second dynamic mode is exponential reduction and is formulated as follows:
\begin{equation}\label{fldr-e}
    \textsc{Reduce}(\kappa, \alpha, \omega, R) = \exp\left(\ln(\kappa) - \frac{\ln(\alpha) - \ln(\omega)}{R}\right).    
\end{equation}
In both functions, $\kappa$ is the compression rate of the previous round, $R$ is the number of global training rounds, $\alpha$ is the initial compression rate, and $\omega$ is the final compression rate. 
Through out $R$ rounds of global training, $\kappa$ is decreased slowly from $\alpha$ to $\omega$, and the $\textsc{Reduce}(\kappa, \alpha, \omega, R)$ function simply computes the next step of $\kappa$ reduction. 

\section{Experimental Evaluation} \label{sec:experiment}




In this section, the proposed FLTS and FLDR algorithms are evaluated.
We utilize CIFAR10, a multi-class classification dataset popular for SNN performance evaluations, to evaluate our proposed compression approaches, including $50,000$ $32\times32$ RGB images of training data and $10,000$ images of validation data from the original dataset. The training data is equally split among clients. If not specified in the experiment, we adopt $5$ clients in the evaluation, and all the clients participate in each global aggregation round. The VGG9 SNN model is applied to the FL framework for evaluation. The sequence length of SNNs, in which each neuron perceives the input data by receiving spikes, will be set to $25$, leveraging the leaky-integrate-and-fire (LIF) model for neuron behavior. The training is conducted with SGD optimizer, and the learning rate is set to $0.1$ with a momentum of $0.95$. The batch size is configured to $32$ across all devices. To better observe the training process, each client performs $1$ local epoch per global round in our experiments, with a total of $100$ global aggregation rounds. 
All the simulation, evaluations of the model and integrated compression algorithms performed on a centralized server with an 80 Core Intel(R) Xeon(R) CPU E5-2698 v4 @ 2.20GHz processor and 503GB of memory. They system also incorporates eight NVIDIA Tesla V100-SXM2-32GB, each with 32 GB of memory. 




\subsection{FL with Top-$\kappa$ Sparsification under Noisy Communication}


\begin{figure}
    \centering
    \subfigure[]{\includegraphics[scale=0.39]{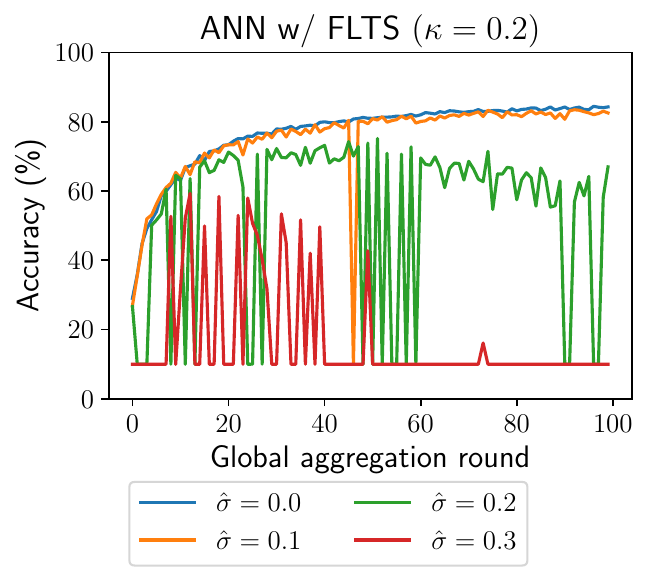}\label{fig:ann_rltv_flts}}
    \subfigure[]{\includegraphics[scale=0.39]{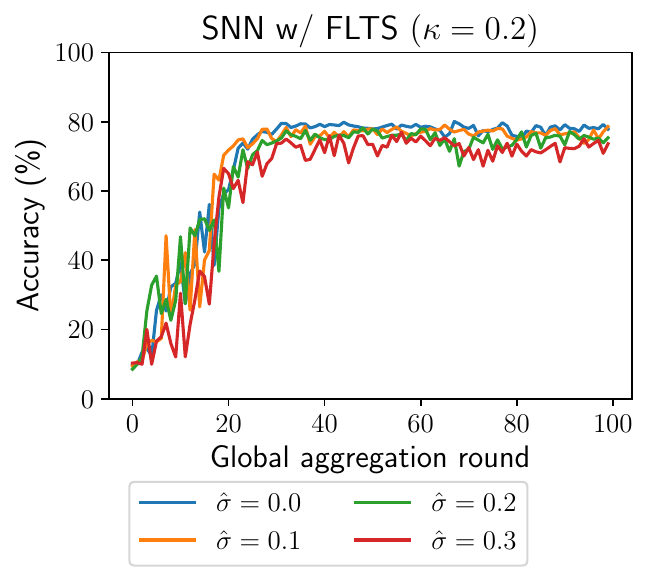}\label{fig:snn_rltv_flts}}
    \caption{Comparison of FL with Top-$\kappa$ Sparification under noisy communication between a) ANNs and b) SNNs.}
    \label{fig:rltv_flts}
\end{figure}

In this section, we conduct experiments on FL with Top-$\kappa$ sparsification (i.e., FLTS) under noisy communication. We aim to compare the performance of ANN and SNN in the setting aforementioned. The compression rate $\kappa$ is set to $0.2$ with various relative noise levels $\hat{\sigma} \in \{0.0, 0.1, 0.2, 0.3\}$. The results are illustrated in Fig. \ref{fig:rltv_flts}. We observe that noisy communication impacts the FL training for both ANN and SNN models. Nevertheless, SNN is significantly more robust than ANN. Specifically, as the noise level increases, the accuracy of ANN becomes much worse, or the model fails to converge, leaving very limited room for communication compression. This phenomenon validates that FL equipped with SNN allows much more bandwidth saving than ANN under practical noisy communication. 

\subsection{Effect of Compression Rate Parameter $\kappa$ in FLTS} 
\label{subsec:flts_prelim}

\begin{figure}
    \centering
    \subfigure[]{\includegraphics[scale=0.4]{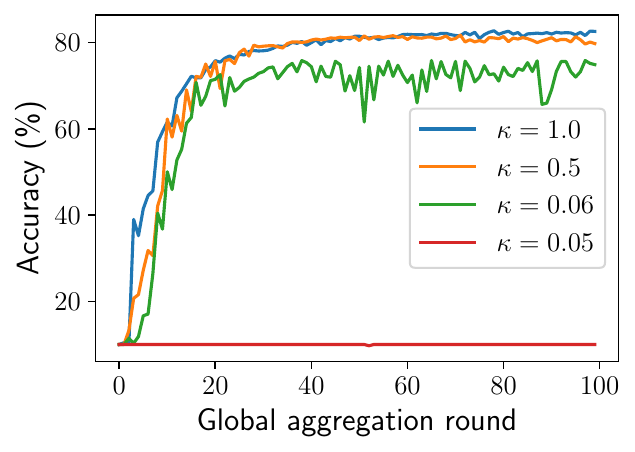}\label{fig:prelim_res_a}}
    \subfigure[]{\includegraphics[scale=0.4]{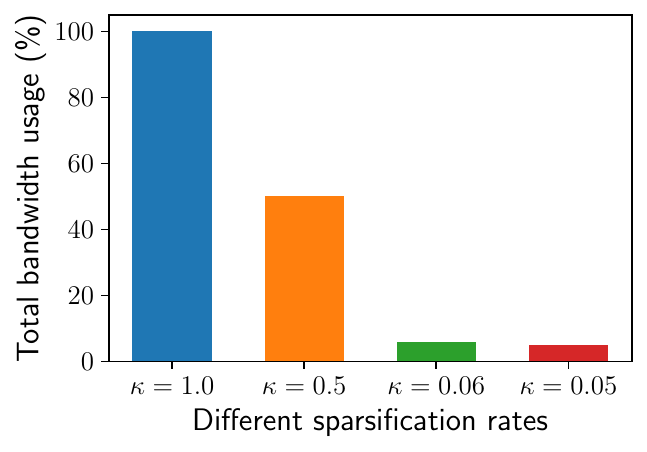}\label{fig:prelim_res_b}}
    \caption{The effectiveness of Top-$\kappa$ sparsification in SNNs by comparing the a) Accuracy, and b) Total bandwidth usage of different sparsification rate $\kappa$. }
    \label{fig:prelim_res}
\end{figure}

This section evaluates the impact of compression rate parameter $\kappa$.
Fig. \ref{fig:prelim_res} presents the training results of FLTS under various $\kappa$ settings such that $\kappa \in \{1.0, 0.5, 0.06, 0.05\}$. 
This evaluation aims to demonstrate the effectiveness of FLTS in terms of accuracy and total bandwidth utilization compared to the baseline without compression, i.e., $\kappa=1.0$, for SNNs. We observe similar accuracy trends with compression rates of $50\%$ ($\kappa = 0.5$) and the baseline without compression ($\kappa = 1.0$). The highest accuracy of $\kappa=1.0$ is $82.7\%$ while $\kappa=0.5$ is $81.7\%$. This result indicates that our FLTS algorithm has great potential to reduce bandwidth overhead in FL with SNNs. Fig. \ref{fig:prelim_res_a} also demonstrates that the FL algorithm normally performs even under $\kappa=0.06$, with the highest accuracy of $75.8\%$, while the FL training does not learn anything under $\kappa=0.05$. Our experimental results indicate that $6\%$ is the minimum compression rate of FLTS for VGG9 SNN with the CIFAR10 dataset. 


\subsection{Impact of Network Size}


\begin{figure}
    \centering
    \subfigure[]{\includegraphics[scale=0.39]{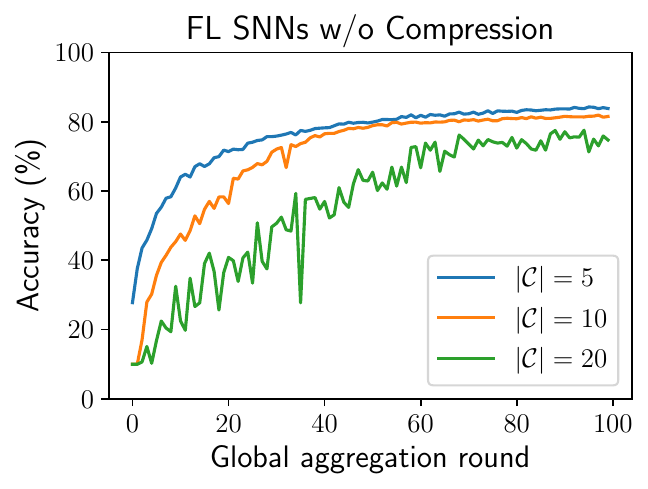}\label{fig:net_size_a}}
    \subfigure[]{\includegraphics[scale=0.39]{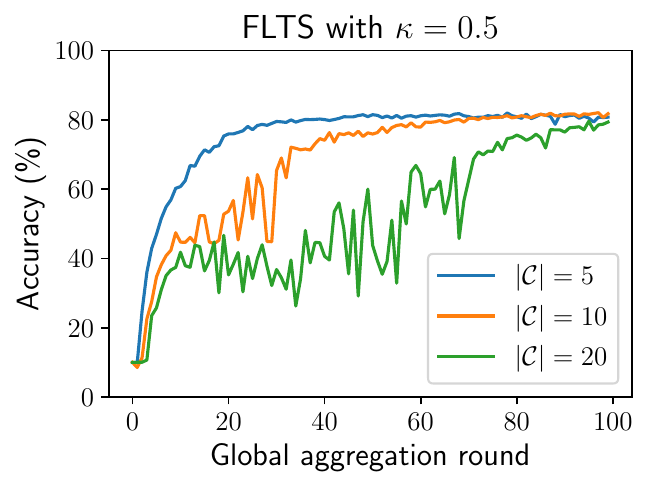}\label{fig:net_size_b}}    
    \caption{Impact of network size, i.e. different number of clients $|\mathcal{C}|$,  on FL with SNNs a) Without compression, and b) FLTS with compression $\kappa = 0.5$.}
    \label{fig:net_size}
\end{figure}

In this section, we investigate the impact of network size. 
Fig. \ref{fig:net_size}(a) and (b) display the impact of the number of clients $(|\mathcal{C}|)$ on the accuracy of FL with SNNs without communication compression, i.e. $\kappa = 1.0$, and FLTS with compression rate $\kappa=0.5$, respectively. 
First, as the number of clients increases, the accuracy curves degrade in uncompressed and compressed scenarios. This observation is expected as it takes more global aggregation rounds in FL to fuse the information in a network with a larger number of clients.  
In addition, our proposed compression method incurs higher variability in accuracy compared to the uncompressed baseline when the network size increases. Specifically, there are more fluctuations, especially in the early stages of the model training. We analyze that this is due to the fact that as the data are spread to more clients, it becomes more difficult for the model to converge. Communication compression could further slow down the training process. Interestingly, we find that the final accuracy of our proposed FLTS algorithm is comparable to or even slightly higher than the baseline with no compression.

\subsection{Effect of the Final Compression Rate $\omega$ in FLDR}
\label{subsec:lin_omega}





\begin{table}[]
\caption{Benchmarking different settings of FLDR: 1) Total Bandwidth consumption (fraction compared to no compression) to attain specific level of Accuracy, and 2) Highest Accuracy achieved}
\scalebox{0.7}{
  
\begin{tabular}{|ll|llllll|}
\hline
\multicolumn{2}{|l|}{}                                                                    & \multicolumn{6}{c|}{\textbf{FLDR}}                                                                                                                                                                                             \\ \cline{3-8} 
\multicolumn{2}{|l|}{}                                                                    & \multicolumn{3}{c|}{\textbf{Linear Reduction FLDR-L}}                                                                                  & \multicolumn{3}{c|}{\textbf{Exponential Reduction FLDR-E}}                                                              \\ \cline{3-8} 
\multicolumn{2}{|l|}{\multirow{-3}{*}{\textbf{Metrics}}}                                  & \multicolumn{1}{l|}{$\omega = 0.01$} & \multicolumn{1}{l|}{$\omega = 0.001$} & \multicolumn{1}{l|}{$\omega = 0.0001$} & \multicolumn{1}{l|}{$\omega = 0.01$} & \multicolumn{1}{l|}{$\omega = 0.001$} & $\omega = 0.0001$       \\ \hline
\multicolumn{1}{|l|}{}                                                       & $25\%$     & \multicolumn{1}{l|}{$0.13$}          & \multicolumn{1}{l|}{$0.12$}           & \multicolumn{1}{l|}{\textbf{0.07}}            & \multicolumn{1}{l|}{$0.1$}           & \multicolumn{1}{l|}{$0.13$}           & $0.09$                  \\ \cline{2-8} 
\multicolumn{1}{|l|}{}                                                       & $40\%$     & \multicolumn{1}{l|}{$0.11$}          & \multicolumn{1}{l|}{$0.1$}            & \multicolumn{1}{l|}{$0.11$}            & \multicolumn{1}{l|}{$0.1$}           & \multicolumn{1}{l|}{$0.1$}            & \textbf{0.08}                  \\ \cline{2-8} 
\multicolumn{1}{|l|}{}                                                       & $50\%$     & \multicolumn{1}{l|}{$0.09$}          & \multicolumn{1}{l|}{$0.08$}           & \multicolumn{1}{l|}{$0.08$}            & \multicolumn{1}{l|}{$0.08$}          & \multicolumn{1}{l|}{$0.07$}           & \textbf{0.06}                  \\ \cline{2-8} 
\multicolumn{1}{|l|}{}                                                       & $60\%$     & \multicolumn{1}{l|}{$0.08$}          & \multicolumn{1}{l|}{$0.09$}           & \multicolumn{1}{l|}{$0.08$}            & \multicolumn{1}{l|}{$0.07$}          & \multicolumn{1}{l|}{$0.07$}           & \textbf{0.06}                  \\ \cline{2-8} 
\multicolumn{1}{|l|}{}                                                       & $70\%$     & \multicolumn{1}{l|}{$0.08$}          & \multicolumn{1}{l|}{$0.07$}           & \multicolumn{1}{l|}{$0.07$}            & \multicolumn{1}{l|}{$0.07$}          & \multicolumn{1}{l|}{$0.07$}           & \textbf{0.05}                  \\ \cline{2-8} 
\multicolumn{1}{|l|}{\multirow{-6}{*}{\begin{turn}{90}\textbf{\begin{tabular}[c]{@{}l@{}}Bandwidth \\ for Accuracy\end{tabular}}\end{turn}}}      & $75\%$     & \multicolumn{1}{l|}{$0.13$}          & \multicolumn{1}{l|}{$0.1$}            & \multicolumn{1}{l|}{$0.1$}             & \multicolumn{1}{l|}{$0.09$}          & \multicolumn{1}{l|}{\textbf{0.06}}           & \cellcolor[HTML]{9B9B9B} \\ \hline
\multicolumn{2}{|l|}{\textbf{\begin{tabular}[c]{@{}l@{}}Highest\\ Accuracy\end{tabular}}} & \multicolumn{1}{l|}{$78.83\%$}                & \multicolumn{1}{l|}{79.65}                 & \multicolumn{1}{l|}{$79.78\%$}                  & \multicolumn{1}{l|}{$78.76\%$}                & \multicolumn{1}{l|}{$77.19\%$}                 &          $73.35\%$               \\ \hline
\end{tabular}
}

\end{table}

This section evaluates the effect of the final compression rate $\omega$ in FLDR. The highest accuracy is obtained by finding the maximum accuracy value within $100$ global aggregation rounds.
The motivation for this experiment is to understand the accuracy pattern when attempting to further reduce bandwidth usage by lowering $\omega$. As observed in TABLE I, we find no significant accuracy difference for linear reduction scheme under various $\omega \in \{0.01, 0.001, 0.0001\}$. Similarly, there is no substantial difference in total bandwidth consumption when varying the $\omega$ value.
We conduct another experiment by varying $\omega$ in FLDR with exponential reduction. Different from FLDR-L, we observe that the accuracy trend of FLDR-E becomes more and more destabilized in exponential reduction as the $\omega$ value declines. The main reason for this observation is that the compression rate decline of the first few rounds in FLDR-E is much steeper compared to that of FLDR-L. Furthermore, we observe that the total bandwidth usage is reduced significantly in lower $\omega$ settings. 
These observations illustrate the capability of FLDR to maintain high model accuracy with significantly lower communication costs, making them promising solutions for FL scenarios where bandwidth availability is extremely limited.

\subsection{Comparison of FLTS and FLDR Algorithms} \label{subsec:different_alg}

\begin{figure}
    \centering
    \subfigure[]{\includegraphics[scale=0.4]{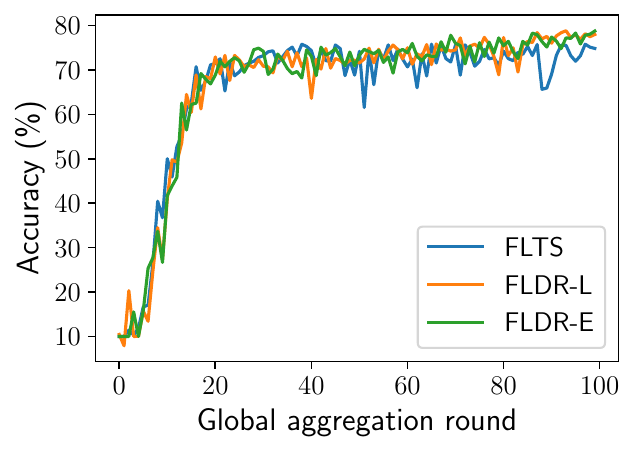}\label{fig:different_alg_a}}
    \subfigure[]{\includegraphics[scale=0.4]{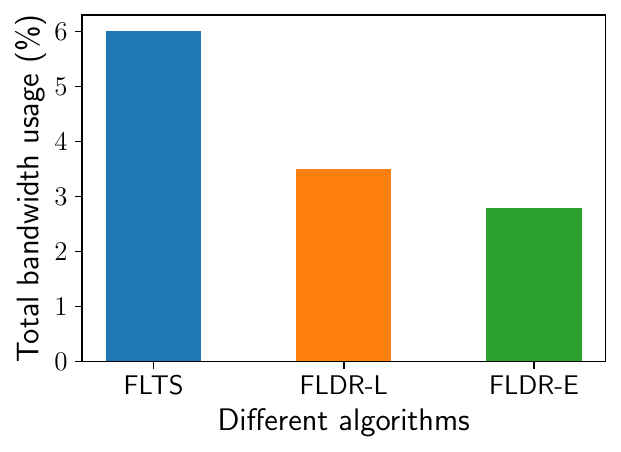}\label{fig:different_alg_b}}
    \caption{Comparison of proposed algorithms: FLTS $\left(\kappa = 0.06\right)$, FLDR-L $\left(\{\alpha=0.06, \omega=0.01\}\right)$ and FLDR-E $\left(\{\alpha=0.06, \omega=0.01\}\right)$.}
    \label{fig:different_alg}
\end{figure}

In this experiment, we compare our proposed FLTS and FLDR algorithms to investigate their characteristics further. For FLDR, two schemes are being evaluated that are based on equations \eqref{fldr-l} and \eqref{fldr-e}, respectively. They are denoted as FLDR-L and FLDR-E. To ensure a fair comparison, $\kappa$ is set to $0.06$ for FLTS, and we apply reduction parameters $\{\alpha=0.06, \omega=0.01\}$ for both FLDR-L and FLDR-E. As shown in Fig. \ref{fig:different_alg_a}, FLDR-L and FLDR-E are slightly better than FLTS, with accuracy values $78.83\%$, $78.76\%$, and $75.8\%$, respectively. Fig. \ref{fig:different_alg_b} summarizes the total bandwidth consumption of three approaches, where FLDR-L ($3.5\%$) and FLDR-E ($2.79\%$) are also better than that of FLTS. Note that the model training fails to converge when we attempt to reduce the total bandwidth in FLTS to $5\%$ in Fig. \ref{fig:prelim_res}. However, Fig. \ref{fig:different_alg} demonstrates that the communication efficiency with FLDR can be further improved in that the parameter reduction is doubled compared to the lowest threshold of FLTS.

\section{Related Work} \label{sec:relate}

\textbf{Robustness of SNN}.
Patel et al. \cite{patel_impact_2023} investigate noise impacts on SNN models when noise is injected into the input and the training process. Kundu et al. \cite{Kundu_HIRE-SNN21} develop an SNN training algorithm that uses crafted input noise in order to harness the model's robustness to gradient-based attacks. Ma et al. \cite{ma_exploiting_2023} further extend this line of efforts by proposing a Noisy Spiking Neural Network (NSNN) paradigm, intentionally incorporating noise to improve the model's robustness against adversarial attacks and challenging perturbations. These studies reveal the effectiveness and resiliency of SNNs to noise added to stages of model training. However, they do not target the FL context as investigated in our work.  

\textbf{FL with SNN}.
Tumpa et al. \cite{tumpa_federated_2023} evaluate the performance of FL with SNN in heterogeneous systems. Following this line of research, recent work in \cite{xie_efficient_2022, aouedi_hfedsnn_2023,liu_energy-efficient_2024} demonstrates the efficiency of SNN with various FL environments and applications. 
However, these works mainly focus on leveraging the energy efficiency of SNN in FL systems. 
Venkatesha et al.~\cite{venkatesha_federated_2021} demonstrate the robustness of SNN in FL applications, including straggling, model dropout, and gradient noise. In this work, we further examine the robustness of SNN and design FL sparsification algorithms for SNN training to reduce communication costs while maintaining model accuracy.

\textbf{Communication Compression in FL}.
Various methods based on compressing the model updates have been proposed to overcome the insufficient bandwidth issues in FL systems \cite{Wang_Fl_compress18,Sattler_FL_compress19}. However, these works are based on the conventional ANN models. In contrast, several works have been conducted to explore the communication efficiency in FL systems with SNNs and provide insights into the trade-offs between communication load and accuracy \cite{skatchkovsky_federated_2020,chaki_communication_2023,liu_federal_2022}. However, they do not consider noisy communication in practical FL systems.

\section{Conclusion} \label{sec:conclude}
In this article, we presented a comprehensive study on the enhancement of communication efficiency in Federated Learning (FL) using Spiking Neural Networks (SNNs).
In light of the inherent robustness of SNNs to noise, we designed Top-$\kappa$ Sparsification based schemes to reduce communication cost in FL without compromising model accuracy. 
The empirical results demonstrated that FL with SNNs saves significantly more bandwidth than their ANNs counterparts under noisy communication. Specifically, under the effects of noise and compression in communication, SNNs can still achieve high model accuracy, while the training for ANNs may fail to converge. 
Our research findings lay the foundation for further exploring the characteristics of SNNs as an alternative to ANNs to improve network efficiency in FL.
\section{Acknowledgments}
This article has been authored by an employee of National Technology \& Engineering Solutions of Sandia, LLC under Contract No.~DE-NA0003525 with the U.S. Department of Energy (DOE). The employee owns all right, title and interest in and to the article and is solely responsible for its contents. The United States Government retains and the publisher, by accepting the article for publication, acknowledges that the United States Government retains a non-exclusive, paid-up, irrevocable, world-wide license to publish or reproduce the published form of this article or allow others to do so, for United States Government purposes. The DOE will provide public access to these results of federally sponsored research in accordance with the DOE Public Access Plan \url{https://www.energy.gov/downloads/doe-public-access-plan}.

\bibliographystyle{IEEEtran}
\bibliography{MAIN}

\end{document}